\documentclass[11pt,letterpaper]{article}
\usepackage[letterpaper]{geometry}
\usepackage{tacl}
\usepackage{times}
\usepackage{latexsym}
\usepackage{amsmath}
\usepackage{multirow}
\usepackage{url}
\makeatletter
\newcommand{\@BIBLABEL}{\@emptybiblabel}
\newcommand{\@emptybiblabel}[1]{}
\makeatother
\usepackage[hidelinks]{hyperref}

\usepackage{latexsym}
\usepackage{amsmath}
\usepackage{graphicx}
\usepackage{mathtools}
\usepackage{amsfonts}
\usepackage{mathrsfs}
\usepackage{multirow}
\usepackage{tablefootnote}

\usepackage{tikz}
\usetikzlibrary{arrows,decorations.pathmorphing,backgrounds,positioning,fit,petri,plotmarks,shapes}
\usetikzlibrary{decorations.markings}

\title{Shortcut Sequence Tagging}

\newcommand*{\affaddr}[1]{#1} 
\newcommand*{\affmark}[1][*]{\textsuperscript{#1}}
\newcommand*{\email}[1]{\texttt{#1}}

\author{%
Huijia Wu\affmark[1,3], Jiajun Zhang\affmark[1,3], and Chengqing Zong\affmark[1,2,3]\\
\affaddr{\affmark[1]National Laboratory of Pattern Recognition, Institute of Automation, CAS}\\
\affaddr{\affmark[2]CAS Center for Excellence in Brain Science and Intelligence Technology }\\
\affaddr{\affmark[3]University of Chinese Academy of Sciences} \\
\email{\{huijia.wu,jjzhang,cqzong\}@nlpr.ia.ac.cn}
}

\date{}

\begin{document}
\maketitle
\begin{abstract}
Deep stacked RNNs are usually hard to train. Adding shortcut connections across different layers is a common way to ease the training of stacked networks. However, extra shortcuts make the recurrent step more complicated. To simply the stacked architecture, we propose a framework called shortcut block, which is a marriage of the gating mechanism and shortcuts, while discarding the self-connected part in LSTM cell. We present extensive empirical experiments showing that this design makes training easy and improves generalization. We propose various shortcut block topologies and compositions to explore its effectiveness. Based on this architecture, we obtain a 6\% relatively improvement over the state-of-the-art on CCGbank supertagging dataset. We also get comparable results on POS tagging task.
\end{abstract}

\section{Introduction}
\label{intro}

In natural language processing, sequence tagging mainly refers to the tasks of assigning discrete labels to each token in a sequence. Typical examples include Part-of-Speech (POS) tagging and Combinatory Category Grammar (CCG) supertagging. A regular feature of sequence tagging is that the input tokens in a sequence cannot be assumed to be independent since the same token in different contexts can be assigned to different tags. Therefore, the classifier should have memories to remember the contexts to make a correct prediction.

Bidirectional LSTMs \cite{graves2005framewise} become dominant in sequence tagging problems due to the superior performance \cite{wang2015part,lample2016neural}. The horizontal hierarchy of LSTMs with bidirectional processing can remember the long-range dependencies without affecting the short-term storage. Although the models have a deep horizontal hierarchy (the depth refers to sequence length), the vertical hierarchy is often shallow, which may not be efficient at representing each token. Stacked LSTMs are deep in both dimensions, but become harder to train due to the feed-forward structure of stacked layers. 

Shortcut connections (shortcuts, or skip connections) enable unimpeded information flow by adding direct connections across different layers \cite{raiko2012deep,graves2013generating,hermans2013training}. Recent works have shown the effectiveness of using shortcuts in deep stacked models \cite{he2015deep,srivastava2015highway,wu2016skip}. These works share a common way of adding shortcuts as increments to the original network. 

We focus on the refinement of shortcut stacked models to make the training easy. Particularly, for stacked LSTMs, the shortcuts make the computation of LSTM blocks more complicated. We replace the self-connected parts in LSTM cells with the shortcuts to simplify the updates. Based on this construction, we introduce the shortcut block, which can be viewed as a marriage of the gating mechanism and the shortcuts, while discard the self-connected units in LSTMs.

Our contribution is mainly in the exploration of shortcut blocks. We propose a framework of stacked LSTMs within shortcuts, using deterministic or stochastic gates to control the shortcut connections. We present extensive experiments on the Combinatory Category Grammar (CCG) supertagging task to compare various shortcut block topologies, gating functions, and combinations of the blocks. We also evaluate our model on Part-of-Speech (POS) tagging task to test the generalization performance. Our model obtains the state-of-the-art results on CCG supertagging and comparable results on POS tagging. 


\section{Recurrent Neural Networks for Sequence Tagging}
Consider a recurrent neural network applied to sequence tagging: Given a sequence $x = (x_1, \ldots, x_T)$, the RNN computes the hidden state $h = (h_1, \ldots, h_T)$ and the output $y = (y_1, \ldots, y_T)$ by iterating the following equations:
\begin{align}
    h_t &= f(x_t, h_{t-1}; \theta_h) \label{rnn_trans} \\
    y_t &= g(h_t; \theta_o)
\end{align}
where $t \in \{1, \ldots, T\}$ represents the time. $x_t$ represents the input at time $t$, $h_{t-1}$ and $h_t$ are the previous and the current hidden state, respectively. $f$ and $g$ are the transition function and the output function, respectively. $\theta_h$ and $\theta_o$ are network parameters.

We use a negative log-likelihood cost to evaluate the performance, which can be written as:
\begin{align}
\mathcal{C} = - \frac{1}{N} \sum_{n=1}^N \log {y}_{t^n}
\end{align}
where $t^n \in \mathbb{N}$ is the true target for sample $n$, and ${y}_{t^n}$ is the $t$-th output in the \textit{softmax} layer given the inputs ${x}^n$.

Stacked RNN is one type of deep RNNs, which refers to the hidden layers are stacked on top of each other, each feeding up to the layer above:
\begin{align}
    h_t^{l} = f^{l}(h_t^{l-1}, h_{t-1}^{l})
\end{align}
where $h_t^{l}$ is the $t$-th hidden state of the $l$-th layer.

\section{Explorations of Shortcuts} \label{skip}
Shortcuts(skip connections) are cross-layer connections, which means the output of layer $l-1$ is not only connected to the layer $l$, but also connected to layer $l+1, \ldots, L$. In this section, we first introduce the traditional stacked LSTMs. Based on this architecture, we propose a shortcut block structure, which is the basic element of our stacked models. 


\subsection{Tranditional Stacked LSTMs}
Stacked LSTMs without skip connections can be defined as:
\begin{align} \label{no_skip}
\begin{split}
    \left(\!
    \begin{array}{c}
      i \\
      f \\
      o \\
      s
    \end{array}
    \!\right) &=
    \left(\!
    \begin{array}{c}
      \text{sigm} \\
      \text{sigm} \\
      \text{sigm} \\
      \text{tanh}
    \end{array}
    \!\right) W^{l} 
    \left(\!
    \begin{array}{c}
    h_t^{l-1} \\
    h_{t-1}^{l}
    \end{array}
    \!\right)
\end{split} \\
    \begin{split}
        c_t^l &= f \odot c_{t-1}^l + i \odot s_t^{l} \\
        h_t^l &= o \odot \text{tanh}(c_t^l)
    \end{split}
\end{align}
During the forward pass, LSTMs need to calculate $c_t^l$ and $h_t^l$, which is the cell's internal state and the cell outputs state, respectively. $c_t^l$ is computed by adding two parts: one is the cell increment $s_t^l$, controlled by the input gate $i_t^{l}$, the other is the self-connected part $c_{t-1}^l$, controlled by the forget gate $f_t^l$. The cell outputs $h_t^l$ are computed by multiplying the activated cell state by the output gate $o_t^l$, which learns when to access memory cell and when to block it. ``$\text{sigm}$'' and ``$\text{tanh}$'' are the sigmoid and tanh activation function, respectively. $W^l \in \mathbb{R}^{4n \times 2n}$ is the weight matrix needs to be learned.

\subsection{Shortcut Blocks}
The hidden units in stacked LSTMs have two forms. One is the hidden units in the same layer $\{h_t^l, t \in 1, \ldots, T\}$, which are connected through an LSTM. The other is the hidden units at the same time step $\{h_t^{l}, l \in 1, \ldots, L\}$, which are connected through a feed-forward network. LSTM can keep the short-term memory for a long time, thus the error signals can be easily passed through $\{1, \ldots, T\}$. However, when the number of stacked layers is large, the feed-forward network will suffer the gradient vanishing/exploding problems, which make the gradients hard to pass through $\{1, \ldots, L\}$. 

Shortcut connections can partly solve the above problem by adding a direct link between layers. An intuitive explanation is that such link can make the error signal passing jump the layers, not just one by one. This may lead to faster convergence and better generalization.  To clarify notations, we introduce the shortcut block, which is composed of the different layers connected through shortcuts. 

Our shortcut block is mainly based on Wu et al. \shortcite{wu2016skip}, which introduce gated shortcuts connected to cell outputs. The main difference is we replace the self-connected parts with shortcuts to compute the internal state. LSTM block \shortcite{hochreiter1997lstm} composes memory cells sharing the same input and output gate. He et al. \shortcite{he2015deep} create a residual block which adds shortcut connections across different CNN layers. All these inspired us to build a shortcut block across different LSTM layers. Our shortcut block is defined as follows:
\begin{align} \label{eq:shortcut1}
\begin{split}
    \left(\!
    \begin{array}{c}
      i \\
      g \\
      o \\
      s
    \end{array}
    \!\right) &=
    \left(\!
    \begin{array}{c}
      \text{sigm} \\
      \text{sigm} \\
      \text{sigm} \\
      \text{tanh}
    \end{array}
    \!\right) W^{l} 
    \left(\!
    \begin{array}{c}
    h_t^{l-1} \\
    h_{t-1}^{l}
    \end{array}
    \!\right)
\end{split} \\
	\begin{split}
    m &=  i \odot s_t^{l} + {g} \odot {h_t^{-l}} \\
    h_t^l &= o \odot \text{tanh}(m) + {g} \odot {h_t^{-l}}
    \end{split}
\end{align}
where $h_t^{-l}$ is the output from one of the previous layers $1, \ldots, l-$1. $g$ is the gate which is used to access the skipped output $h_t^{-l}$ or block it. 

\paragraph{Comparison with LSTMs.}
LSTMs introduce a memory cell with a fixed self-connection to make the constant error flow.    
They compute the following increment to the self-connected cell at each time step:
\begin{equation}
    c_t = c_{t-1} + s_t
\end{equation}
Here we remove the multiplicative gates to simplify the explanation. The self-connected cell $c_t$ can keep the recurrent information for a long time. $s_t$ is the increment to the cell. While in our shortcut block, we only consider the increment part. Since in sequence tagging problem, the input sequence and the output sequence are exactly match. Specifically, the input token $x_i, i \in \{1, \ldots, n\}$ in a input sequence with length $n$ provides the most relevant information to predict the corresponding label $y_i, i \in \{1, \ldots, n\}$ in a output sequence. We want to focus the information flow in the vertical direction through shortcuts, rather than in the horizontal direction through the self-connected units. Therefore, we only consider the increment to the cell and ignore the self-connected part. Our cell state becomes:
\begin{equation}
    m = h_t^{-l} + s_t
\end{equation}

\paragraph{Why Only Increments?}
Obviously, we can keep the self-connected part with the increments in the cell state. But the most important reason to this design is that it is much easier to compute. We do not need extra space to preserve the cell state. This makes deep stacked models much easier to train.

\paragraph{Discussion.}
The shortcut block can be seen as a generalization of several multiscale RNN architectures. It is up to the user to define the block topology. For example, when $h_t^{-l} := h_t^{l-1}$, this is similar to recurrent highway networks \cite{zilly2016recurrent} and highway LSTMs \cite{zhang2016highway}. When $h_t^{-l} := h_t^{l-2}$, this becomes the traditional shortcuts for RNNs:
\begin{align} \label{eq:shortcut2}
\begin{aligned}
    m &= i \odot s_t^{l} + {g} \odot {h_t^{l-2}}\\
    h_t^l &= o \odot \text{tanh}(m) + {g} \odot {h_t^{l-2}}
\end{aligned}
\end{align}


\subsection{Gates Sesign} \label{sec:compute}
Shortcut gates are used to make the skipped path deterministic \cite{srivastava2015highway} or stochastic \cite{huang2016deep}. We explore many ways to compute the shortcut gates (denoted by $g_t^l$). The simplest case is to use $g_t^l$ as a linear operator. In this case, $g_t^l$ is a weight matrix, and the element-wise product $g_t^l \odot h_t^{-l}$ in Eq. \eqref{eq:shortcut1} becomes a matrix-vector multiplication:
\begin{align} \label{eq:gate1}
g_t^l \odot h_t^{-l} & := W^l h_t^{-l}
\end{align}

We can also get $g_t^l$ under a non-linear mapping, which is similar to the computation of gates in LSTM:
\begin{align} \label{eq:gate2}
g_t^{l} &= \sigma(W^l h_{t}^{l-1})
\end{align}
Here we use the output of layer $l-$1 to control the shortcuts, e.g. $h_t^{l-2}$. Notice that this non-linear mapping is not unique, we just show the simplest case.

Furthermore, inspired by the dropout \cite{srivastava2014dropout} strategy, we can sample from a Bernoulli stochastic variable to get $g_t^l$. In this case, a deterministic gate is transformed into a stochastic gate. 
\begin{align} \label{eq:gate3}
g_t^l \sim \text{Bernoulli}(p)
\end{align}
where $g_t^l$ is a vector of independent Bernoulli random variables each of which has probability $p$ of being 1. We can either fix $p$ with a specific value or learn it with a non-linear mapping. For example, we can learn $p$ by:
\begin{align}
p = \sigma(H^l h_t^{l-1})
\end{align}
At test time, $h_t^{-l}$ is multiplied by $p$. 

\paragraph{Discussion.}
The gates of LSTMs are essential parts to avoid weight update conflicts, which are also invoked by the shortcuts. In experiments, we find that using deterministic gates is better than the stochastic gates. We recommend using the logistic gates to compute $g_t^l$.

\subsection{Compositions of Shortcut Blocks}
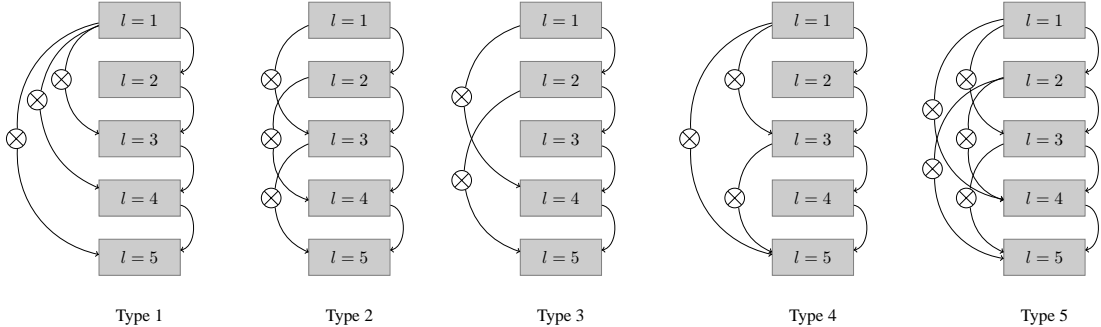
\begin{figure*}
\centering
\scalebox{.6}{
\hspace{5mm}
\begin{minipage}[b]{0.3\textwidth}
\begin{tikzpicture}
[node/.style={rectangle,draw=black!50,fill=black!20,
             inner sep=0pt,minimum width=1.8cm,minimum height = 0.8cm},
gate/.style={circle,draw=black,scale=1.2,path picture={%
        \draw[black]
         (path picture bounding box.south west) -- (path picture bounding box.north east) 
         (path picture bounding box.north west) -- (path picture bounding box.south east);
        }},
caption/.style={rectangle,draw=white,inner sep=0pt,
				minimum width=1.8cm,minimum height = 0.8cm}]
\node[node] (node1) {$l=1$};
\node[node] (node2) [below=0.5cm of node1] {$l=2$};
\node[node] (node3) [below=0.5cm of node2]{$l=3$};
\node[node] (node4) [below=0.5cm of node3]{$l=4$};
\node[node] (node5) [below=0.5cm of node4]{$l=5$};
\node[caption] (node6) [below=0.5cm of node5]{Type 1};
\node[gate] (gate1) [left=0.6cm of node2]{};
\node[gate] (gate2) [left=1.6cm of node3]{};
\node[gate] (gate3) [below left=1.7cm of node1]{};

\draw [->] (node1)        to [bend left=80] (node2);
\draw [->] (node2)        to [bend left=80] (node3);
\draw [->] (node3)        to [bend left=80] (node4);
\draw [->] (node4)        to [bend left=80] (node5);

\draw [-] (node1)        to [bend right=30] (gate1);
\draw [->] (gate1) to [bend right=30] (node3);
\draw [-] (node1)        to [bend right=40] (gate2);
\draw [->] (gate2) to [bend right=40] (node5);
\draw [-] (node1)        to [bend right=30] (gate3);
\draw [->] (gate3) to [bend right=30] (node4);
\end{tikzpicture}
\end{minipage}
\hspace{5mm}
\begin{minipage}[b]{0.25\textwidth}
\begin{tikzpicture}
[node/.style={rectangle,draw=black!50,fill=black!20,
             inner sep=0pt,minimum width=1.8cm,minimum height = 0.8cm},
gate/.style={circle,draw=black,scale=1.2,path picture={%
        \draw[black]
         (path picture bounding box.south west) -- (path picture bounding box.north east) 
         (path picture bounding box.north west) -- (path picture bounding box.south east);
        }},
caption/.style={rectangle,draw=white,inner sep=0pt,
				minimum width=1.8cm,minimum height = 0.8cm}]]
\node[node] (node1) {$l=1$};
\node[node] (node2) [below=0.5cm of node1] {$l=2$};
\node[node] (node3) [below=0.5cm of node2]{$l=3$};
\node[node] (node4) [below=0.5cm of node3]{$l=4$};
\node[node] (node5) [below=0.5cm of node4]{$l=5$};
\node[caption] (node6) [below=0.5cm of node5]{Type 2};
\node[gate] (gate1) [left=0.6cm of node2]{};
\node[gate] (gate2) [left=0.6cm of node3]{};
\node[gate] (gate3) [left=0.6cm of node4]{};

\draw [->] (node1)        to [bend left=80] (node2);
\draw [->] (node2)        to [bend left=80] (node3);
\draw [->] (node3)        to [bend left=80] (node4);
\draw [->] (node4)        to [bend left=80] (node5);

\draw [-] (node1)        to [bend right=30] (gate1);
\draw [->] (gate1) to [bend right=30] (node3);
\draw [-] (node2)        to [bend right=40] (gate2);
\draw [->] (gate2) to [bend right=40] (node4);
\draw [-] (node3)        to [bend right=30] (gate3);
\draw [->] (gate3) to [bend right=30] (node5);
\end{tikzpicture}
\end{minipage}

\begin{minipage}[b]{0.3\textwidth}
\begin{tikzpicture}
[node/.style={rectangle,draw=black!50,fill=black!20,
             inner sep=0pt,minimum width=1.8cm,minimum height = 0.8cm},
gate/.style={circle,draw=black,scale=1.2,path picture={%
        \draw[black]
         (path picture bounding box.south west) -- (path picture bounding box.north east) 
         (path picture bounding box.north west) -- (path picture bounding box.south east);
        }},
caption/.style={rectangle,draw=white,inner sep=0pt,
				minimum width=1.8cm,minimum height = 0.8cm}]]
\node[node] (node1) {$l=1$};
\node[node] (node2) [below=0.5cm of node1] {$l=2$};
\node[node] (node3) [below=0.5cm of node2]{$l=3$};
\node[node] (node4) [below=0.5cm of node3]{$l=4$};
\node[node] (node5) [below=0.5cm of node4]{$l=5$};
\node[caption] (node6) [below=0.5cm of node5]{Type 3};
\node[gate] (gate1) [below left=1.6cm of node1]{};
\node[gate] (gate2) [below=1.4cm of gate1]{};

\draw [->] (node1)        to [bend left=80] (node2);
\draw [->] (node2)        to [bend left=80] (node3);
\draw [->] (node3)        to [bend left=80] (node4);
\draw [->] (node4)        to [bend left=80] (node5);

\draw [-] (node1)        to [bend right=30] (gate1);
\draw [->] (gate1) to [bend right=30] (node4);
\draw [-] (node2)        to [bend right=30] (gate2);
\draw [->] (gate2) to [bend right=30] (node5);
\end{tikzpicture}
\end{minipage}

\begin{minipage}[b]{0.32\textwidth}
\begin{tikzpicture}
[node/.style={rectangle,draw=black!50,fill=black!20,
             inner sep=0pt,minimum width=1.8cm,minimum height = 0.8cm},
gate/.style={circle,draw=black,scale=1.2,path picture={%
        \draw[black]
         (path picture bounding box.south west) -- (path picture bounding box.north east) 
         (path picture bounding box.north west) -- (path picture bounding box.south east);
        }},
caption/.style={rectangle,draw=white,inner sep=0pt,
				minimum width=1.8cm,minimum height = 0.8cm}]]
\node[node] (node1) {$l=1$};
\node[node] (node2) [below=0.5cm of node1] {$l=2$};
\node[node] (node3) [below=0.5cm of node2]{$l=3$};
\node[node] (node4) [below=0.5cm of node3]{$l=4$};
\node[node] (node5) [below=0.5cm of node4]{$l=5$};
\node[caption] (node6) [below=0.5cm of node5]{Type 4};
\node[gate] (gate1) [left=0.6cm of node2]{};
\node[gate] (gate2) [left=1.6cm of node3]{};
\node[gate] (gate3) [left=0.6cm of node4]{};

\draw [->] (node1)        to [bend left=80] (node2);
\draw [->] (node2)        to [bend left=80] (node3);
\draw [->] (node3)        to [bend left=80] (node4);
\draw [->] (node4)        to [bend left=80] (node5);

\draw [-] (node1)        to [bend right=30] (gate1);
\draw [->] (gate1) to [bend right=30] (node3);
\draw [-] (node1)        to [bend right=40] (gate2);
\draw [->] (gate2) to [bend right=40] (node5);
\draw [-] (node3)        to [bend right=30] (gate3);
\draw [->] (gate3) to [bend right=30] (node5);
\end{tikzpicture}
\end{minipage}
\begin{minipage}[b]{0.35\textwidth}
\begin{tikzpicture}
[node/.style={rectangle,draw=black!50,fill=black!20,
             inner sep=0pt,minimum width=1.8cm,minimum height = 0.8cm},
gate/.style={circle,draw=black,scale=1.2,path picture={%
        \draw[black]
         (path picture bounding box.south west) -- (path picture bounding box.north east) 
         (path picture bounding box.north west) -- (path picture bounding box.south east);
        }},
caption/.style={rectangle,draw=white,inner sep=0pt,
				minimum width=1.8cm,minimum height = 0.8cm}]]
\node[node] (node1) {$l=1$};
\node[node] (node2) [below=0.5cm of node1] {$l=2$};
\node[node] (node3) [below=0.5cm of node2]{$l=3$};
\node[node] (node4) [below=0.5cm of node3]{$l=4$};
\node[node] (node5) [below=0.5cm of node4]{$l=5$};
\node[caption] (node6) [below=0.5cm of node5]{Type 5};
\node[gate] (gate1) [left=0.6cm of node2]{};
\node[gate] (gate3) [left=0.6cm of node4]{};
\node[gate] (gate4) [left=0.6cm of node3]{};
\node[gate] (gate5) [below left=2cm of node1]{};
\node[gate] (gate6) [below left=2cm of node2]{};

\draw [->] (node1)        to [bend left=80] (node2);
\draw [->] (node2)        to [bend left=80] (node3);
\draw [->] (node3)        to [bend left=80] (node4);
\draw [->] (node4)        to [bend left=80] (node5);

\draw [-] (node1)        to [bend right=30] (gate1);
\draw [->] (gate1) to [bend right=30] (node3);
\draw [-] (node3)        to [bend right=30] (gate3);
\draw [->] (gate3) to [bend right=30] (node5);
\draw [-] (node2)        to [bend right=40] (gate4);
\draw [->] (gate4) to [bend right=40] (node4);
\draw [-] (node1)        to [bend right=40] (gate5);
\draw [->] (gate5) to [bend right=40] (node4);
\draw [-] (node2)        to [bend right=40] (gate6);
\draw [->] (gate6) to [bend right=40] (node5);
\end{tikzpicture}
\end{minipage}%
}
\caption{Compositions of shortcut blocks. We call a shortcut block with span 1 when $h_t^{-l} := h_t^{l-2}$. \label{fig:comp}}
\end{figure*}

In the previous subsections, we introduce the shortcut blocks and the computation of gating functions. To build deep stacked models we need to compose these blocks together. In this section, we discuss several kinds of compositions, as shown in Figure \ref{fig:comp}. The links with $\odot$ represent the gated identity connections across layers. There are two types of connections in Figure \ref{fig:comp}: one is the direct connections between adjacent layers, the other is the gated connections across different layers. 

In Figure \ref{fig:comp}, Type 1 is to connect the input of the first hidden layer $h_t^1$ to all the following layers $L= 2, 3, 4, 5$. Type 2 and Type 3 are composed of the shortcut blocks with span 1 and 2, respectively. Type 4 and Type 5 are the nested shortcut blocks. We wish to find an optimal composition to pass information in deep stacked models. 

\section{Neural Architecture for sequence Tagging}
Sequence tagging can be formulated as $P({t} | {w}; {\theta})$, where ${w} = [w_1, \ldots, w_T]$ indicates the $T$ words in a sentence, and ${t} = [t_1, \ldots, t_T]$ indicates the corresponding $T$ tags. In this section we introduce an neural architecture for $P(\cdot)$, which includes an input layer, a stacked hidden layers and an output layer. Since the stacked hidden layers have already been introduced in the previous section, we only introduce the input and the output layer here.

\subsection{Network Inputs} \label{sec:input}
Network inputs are the representation of each token in a sequence. There are many kinds of token representations, such as using a single word embedding, using a local window approach, or a combination of word and character-level representation. Following Wu et al. \shortcite{wu2016skip}, we use a local window approach together with a concatenation of word representations, character representations, and capitalization representations.

Formally, we can represent the distributed word feature ${f}_{w_t}$ using a concatenation of these embeddings:
\begin{align}
    {f}_{w_t} = [{L}_w(w_t); {L}_a(a_t); {L}_c({c}_w)]
\end{align}
where $w_t$, $a_t$ represent the current word and its capitalization. ${c}_w := [c_1, c_2, \ldots, c_{T_w}]$, where $T_w$ is the length of the word and $c_i, i \in \{1, \ldots, T_w\}$ is the $i$-th character for the particular word.  ${L}_w(\cdot) \in \mathbb{R}^{|V_w|\times n}$, ${L}_a(\cdot) \in \mathbb{R}^{|V_a|\times m}$ and ${L}_c(\cdot) \in \mathbb{R}^{|V_c| \times r}$ are the look-up tables for the words, capitalization and characters, respectively. ${f}_{w_t} \in \mathbb{R}^{n+m+r}$ represents the distributed feature of $w_t$. A context window of size $d$ surrounding the current word is used as an input:
\begin{align} \label{eq:input}
    {x}_t = [{f}_{w_{t-\lfloor d/2 \rfloor}}; \ldots; {f}_{w_{t+\lfloor d/2 \rfloor}}]
\end{align}
where ${x}_t \in \mathbb{R}^{(n+m+r)\times d}$ is a concatenation of the context features. In the following we discuss the $w_t$, $a_t$ and $c_w$ in detail. 

\paragraph{Word Representations.}
All words in the vocabulary share a common look-up table, which is initialized with random initializations or pre-trained embeddings. Each word in a sentence can be mapped to an embedding vector $w_t$. The whole sentence is then represented by a matrix with columns vector $[w_1, w_2, \ldots, w_T]$. Following Wu et al. \shortcite{wu2016dynamic}, we use a context window of size $d$ surrounding with a word $w_t$ to get its context information. Rather, we add logistic gates to each token in the context window. The word representation is computed as $w_t = [r_{{t-\lfloor d/2 \rfloor}} w_{t-\lfloor d/2 \rfloor}; \ldots; r_{{t+\lfloor d/2 \rfloor}} w_{t+\lfloor d/2 \rfloor}]$, where $r_t := [r_{{t-\lfloor d/2 \rfloor}}, \ldots, r_{{t+\lfloor d/2 \rfloor}}] \in \mathbb{R}^{d}$ is a logistic gate to filter the unnecessary contexts, $w_{t-\lfloor d/2 \rfloor}, \ldots, w_{t+\lfloor d/2 \rfloor}$ is the word embeddings in the local window.

\paragraph{Capitalization Representations.}
We lowercase the words to decrease the size of word vocabulary to reduce sparsity, but we need an extra capitalization embeddings to store the capitalization features, which represent whether or not a word is capitalized.

\paragraph{Character Representations.} We concatenate character embeddings in a word to get the character-level representation. Concretely, given a word $w$ consisting of a sequence of characters $[c_1, c_2, \ldots, c_{l_w}]$, where $l_w$ is the length of the word and $L(\cdot)$ is the look-up table for characters. We concatenate the leftmost most 5 character embeddings $L(c_1), \ldots, L(c_5)$ with its rightmost 5 character embeddings $L(c_{l_w-4}), \ldots, L(c_{l_w})$ to get $c_w$. When a word is less than five characters, we pad the remaining characters with the same special symbol. 


\subsection{Network Outputs}
For sequence tagging, we use a \emph{softmax} activation function $g(\cdot)$ in the output layer:
\begin{align}
{y}_t &= g({W^{hy}} [\overrightarrow{h_t}; \overleftarrow{h_t}])
\end{align}
where ${y}_t$ is a probability distribution over all possible tags. $y_k(t) = \frac{\exp(h_k)}{\sum_{k'} \exp(h_{k'})}$ is the $k$-th dimension of ${y}_t$, which corresponds to the $k$-th tag in the tag set. ${W^{hy}}$ is the hidden-to-output weight.

\section{Experiments}
\subsection{Combinatory Category Grammar Supertagging}
Combinatory Category Grammar (CCG) supertagging is a sequence tagging problem in natural language processing. The task is to assign supertags to each word in a sentence. In CCG the supertags stand for the lexical categories, which are composed of the basic categories such as $N$, $NP$ and $PP$, and complex categories, which are the combination of the basic categories based on a set of rules. Detailed explanations of CCG refer to \cite{steedman2000syntactic,steedman2011combinatory}. 

Another similar task is POS tagging, in which the tags are part of speeches. Technically, the two kinds of tags classify the words in different ways: CCG tags implicate the semantics of words, while the POS tags represent the syntax of words. Although these distinctions are important in linguistics, here we all treat them as the activations of neurons, using distributed representations to encode these tags. This high-level abstraction greatly improves the generalization, and heavily reduces the cost of the model redesign. 

\subsubsection{Dataset and Pre-processing}
Our experiments are performed on CCGBank \cite{hockenmaier2007ccgbank}, which is a translation from Penn Treebank \cite{marcus1993building} to CCG with a coverage 99.4\%. We follow the standard splits, using sections 02-21 for training, section 00 for development and section 23 for the test. We use a full category set containing 1285 tags. All digits are mapped into the same digit `9', and all words are lowercased.

\subsubsection{Network Configuration}
\paragraph{Initialization.}
There are two types of weights in our experiments: recurrent and non-recurrent weights. For non-recurrent weights, we initialize word embeddings with the pre-trained 100-dimensional GolVe vectors \cite{pennington2014glove}. Other weights are initialized with the Gaussian distribution $\mathcal{N}(0, \frac{1}{\sqrt{\text{fan-in}}})$ scaled by a factor of 0.1, where \textit{fan-in} is the number of units in the input layer. For recurrent weight matrices, following \cite{saxe2013exact} we initialize with random orthogonal matrices through SVD to avoid unstable gradients. All bias terms are initialized with zero vectors.

\paragraph{Hyperparameters.}
Our context window size is set to 3. The dimension of character embedding and capitalization embeddings are 5. The size of the input layer after concatenation is 465 ((word embedding 100 + cap embedding 5 + character embedding 50) $\times$ window size 3). The number of cells of the stacked bidirectional LSTM is also set to 465 for orthogonal initialization. All stacked hidden layers have the same number of cells. The output layer has 1286 neurons, which equals to the number of tags in the training set with a \textsc{rare} symbol. 

\paragraph{Training.} 
We train the networks using the back-propagation algorithm, using stochastic gradient descent (SGD) algorithm with an initial learning rate 0.02. The learning rate is then scaled by 0.5 when the following condition satisfied: 
\begin{align*}
\frac{|e_p - e_c|}{e_p} <= 0.005 \text{  and  } lr >= 0.0005
\end{align*}
where $e_p$ is the error rate on the validation set on the previous epoch. $e_c$ is the error rate on the current epoch. The explanation of the rule is when the growth of the performance become lower, we need to use a smaller learning rate to adjust the weights. We use on-line learning in our experiments, which means the parameters will be updated on every training sequences, one at a time.

\paragraph{Regularization.}
Dropout \cite{srivastava2014dropout} is the only regularizer in our model to avoid overfitting. Other regularization methods such as weight decay and batch normalization do not work in our experiments. We add a binary dropout mask to the local context windows with a drop rate $p$ of 0.25. We also apply dropout to the output of the first hidden layer and the last hidden layer, with a 0.5 drop rate. At test time, weights are scaled with a factor $1-p$. 

\subsubsection{Comparison with Other Systems}
Table \ref{tab:one} shows the comparison with other models for supertagging. The comparison does not include any externally labeled data or POS tags. We evaluate the models composed of shortcut blocks with different depths. We present experiments trained on the training set and evaluated on the test set using the highest 1-best supertagging accuracy on the development set. 

Our 9-stacked model presents state-of-the-art results (94.99 on test set) comparing with other systems. Notice that 9 is the number of stacked Bi-LSTM layers. The total layer of the networks contains 11 (9 + 1 input-to-hidden layer + 1 hidden-to-output layer) layers. We find the network with stacked depth 7 or 9 achieves better performance than depth 11 or 13, but the difference is tiny. Our stacked models follow Eq. \eqref{eq:shortcut2} and gating functions refer to Eq. \eqref{eq:gate2}.

\begin{table}
\centering
\scalebox{0.85}{
\begin{tabular}{ l|l|l }
\hline \bf Model & \bf Dev & \bf Test \\ \hline \hline
Clark and Curran \shortcite{clark2007wide} & 91.5 & 92.0 \\
MLP (Lewis et al. \shortcite{lewis2014improved}) & 91.3 & 91.6 \\
Bi-LSTM (Lewis et al. \shortcite{lewis2016lstm}) & 94.1 & 94.3 \\
Elman-RNN (Xu et al. \shortcite{xu2015ccg}) & 93.1 & 93.0 \\
Bi-RNN (Xu et al. \shortcite{xu2016expected}) & 93.49 & 93.52 \\
Bi-LSTM (Vaswani et al. \shortcite{vaswani2016supertagging}) & 94.24 & 94.5 \\ 
9-stacked Bi-LSTM (Wu et al. \shortcite{wu2016skip}) & 94.55 & 94.69 \\
\hline
7-stacked shortcut block (Ours) & 94.74 & 94.95 \\
9-stacked: shortcut block (Ours) & \bf 94.82 & \bf 94.99 \\
11-stacked: shortcut block (Ours) & 94.66 & 94.86 \\
13-stacked: shortcut block (Ours) & 94.73 & 94.97  \\
\hline
\end{tabular}
}
\caption{1-best supertagging accuracy on CCGbank\label{tab:one}}
\end{table}


\subsubsection{Exploration of Shortcuts}
To get a better understanding of the shortcut architecture proposed in Eq. \eqref{eq:shortcut1}, we experiment with its variants to compare the performance. Our analysis mainly focuses on three parts: the topology of shortcut blocks, the gating mechanism, and their compositions. The default number of the stacked layers is 7.  We also use the shared gates and Type 2's architecture as our default configurations, which are described in Eq. \eqref{eq:shortcut2}. The comparison is summarized as follows:

\begin{table*}
\centering
\scalebox{0.85}{
\begin{tabular}{ l|l|l|l}
\hline 
\bf Case & \bf Variant & \bf Dev & \bf Test \\ 
\hline \hline
\multirow{1}{*}{$h_t^l$ updated \cite{wu2016skip}}
& with gate: $h_t^l = \tilde{h}_t^l + g \odot h_t^{l-2}$ & 94.51 & 94.67 \\ 
\hline
\multirow{6}{*}{both $c_t^l$ and $h_t^l$ updated (Case 1)} & 
no gate: $\begin{array} {lcl} c_t^l = \tilde{c}_t^l + h_t^{l-2}, h_t^l = \tilde{h}_t^l + h_t^{l-2} \end{array}$ & 93.84 & 93.84 \\
& with gate: $\begin{array} {lcl} c_t^l = \tilde{c}_t^l + g \odot h_t^{l-2}, h_t^l = \tilde{h}_t^l + g \odot h_t^{l-2} \end{array}$ & 94.72 & \bf 95.08 \\
& highway gate: $\begin{array} {lcl} c_t^l = (1 - g) \odot \tilde{c}_t^l + g \odot h_t^{l-2} \\ h_t^l = (1 - g) \odot \tilde{h}_t^l + g \odot h_t^{l-2} \end{array}$ & 94.49 & 94.62 
\\
& shortcuts for both $c_t^l$ and $h_t^l$: $\begin{array} {lcl} c_t^l = \tilde{c}_t^l + g_c \odot c_t^{l-2} \\ h_t^l = \tilde{h}_t^l + g_h \odot h_t^{l-2} \end{array}$ & 94.72 & 94.98
\\ 
\hline
\multirow{5}{*}{shortcut block (Case 2)}
& no gate in $h_t^l$: $h_t^l = o \odot \tanh(m) + h_t^{l-2}$ & 94.15 & 94.29 \\
& no gate in $m$: $m = i \odot s_t^l + h_t^{l-2}$ & \bf 94.77 & 94.97 \\
& share gate: $h_t^l = o \odot \tanh(m) + o \odot h_t^{l-2}$ & 94.68 & 94.83 \\
& no shortcut in internal: $h_t^l = o \odot \tanh(i_t \odot s_t) + g \odot h_t^{l-2}$ & 93.83 & 94.01 \\
& no shortcut in cell output: $h_t^l = o \odot \tanh(m)$ & 93.58 & 93.82 \\ 
\hline
\end{tabular}
}
\caption{Comparsion of shortcut topologies. We use $\tilde{h}_t^l$ to represent the original cell output of LSTM block, which equals $o \odot \text{tanh}(c_t^l)$, similar to $\tilde{c}_t^l := i \odot s_t^l + f \odot c_{t-1}$. \label{tab:two}} 
\end{table*}


\paragraph{Shortcut Topologies.}
Table \ref{tab:two} shows the comparison of shortcut topologies. Here we design two kinds of models for comparison: one is both $c_t^l$ and $h_t^l$ connected through shortcuts (Case 1), the other is using $m$ to replace $c_t^l$ (Case 2), as defined in the shortcut block. We find the skip connections to both the internal states and the cell outputs with multiplicative gating achieves the highest accuracy (case 1, 95.08\%) on the test set. But case 2 can get a better validation accuracy (94.77\%). We prefer to use case 2 since it generalizes well and much easier to train. 

\subsubsection{Comparison of Gating Functions}
We experiment with several gating functions proposed in Section \ref{sec:compute}. Detailed discussions are described below.

\paragraph{Identity Mapping.}
We use the $\tanh$ function to the previous outputs to break the identity link. The result is 94.81\% (Table \ref{tab:three}), which is poorer than the identity function. We can infer that the identity function is more suitable than other scaled functions such as sigmoid or tanh to transmit information.

\paragraph{Exclusive Gating.}
We find deterministic gates performs better than stochastic gates. Further, non-linear mapping $g_t^{l} = \sigma(W^l h_{t}^{l-1})$ achieves the best test accuracy (Table \ref{tab:three}, 94.79\%), while other types such as linear or stochastic gates are not generalize well.

\begin{table*}
\centering
\scalebox{0.8}{
\begin{tabular}{ l|l|l|l}
\hline 
\bf Case & \bf Variant & \bf Dev & \bf Test \\ 
\hline \hline
scaled mapping & replace $h_t^{l-2}$ with $\tanh(h_t^{l-2})$ & 94.60 & 94.81 \\
\hline
\multirow{1}{*}{linear mapping}
& $g_t^l \odot h_t^{-l} = w^l \odot h_t^{-l}$ & 92.07 & 92.15 \\ 
\hline
\multirow{3}{*}{non-linear mapping}
& $g_l^l = \sigma(W^l h_t^{l-1})$ & \bf 94.79 & \bf 94.91 \\
& $g_l^l = \sigma(U^l h_{t-1}^l)$ & 94.21 & 94.56 \\
& $g_l^l = \sigma(V^l h_{t}^{l-2})$ & 94.60 & 94.78 \\
\hline
\multirow{2}{*}{stochastic sampling}
& $g_t^l \sim \text{Bernoulli}(p)$, $p=0.5$ & 91.12 & 91.47 \\
& $g_t^l \sim \text{Bernoulli}(p)$, $p=\sigma({H^l h_t^{l-1}})$ & 93.90 & 94.06 \\
\hline
\end{tabular}
}
\caption{Comparsion of gating functions. The non-linear mapping $g_l^l = \sigma(W^l h_t^{l-1})$ is the preferred choice. \label{tab:three}}
\end{table*}

\subsubsection{Comparison of Shortcut Block Compositions}
We experiment with several kinds of compositions of shortcut blocks, as shown in Table \ref{tab:five}. We find that shortcut block with span 1 (Type 2 and 5) perform better than other spans (Type 1, 3 and 4). In experiments, we use Type 2 as our default configuration since it is much easier to compute than Type 5.
\begin{table}
\centering
\scalebox{0.9}{
\begin{tabular}{ c|l|l}
\hline 
\bf Type & \bf Dev & \bf Test \\ 
\hline \hline
1 & 94.22 & 94.38 \\
\hline
2 & 94.79 & 94.94 \\
\hline 
3 & 94.53 & 94.80 \\
\hline 
4 & 94.55 & 94.70 \\
\hline
5 & 94.76 & 94.95 \\
\hline 
\end{tabular}
}
\caption{Comparsion of shortcut block combinations. Dense compositions (Type 2 and 5) performs better than sparse ones. \label{tab:five}}
\end{table}

\subsubsection{Comparison of Hyper-parameters}
As described in Section \ref{sec:input}, we use a complex input encoding for our model. Concretely, we use a context window approach, together with character-level information to get a better representation for the raw input. We give comparisons for the system with/without this approaches while keeping the hidden and the output parts unchanged. 

Table \ref{tab:four} shows the effects of the hyper-parameters on the task. We find that the model does not perform well (94.06\%) without using local context windows. Although LSTMs can memorize recent inputs for a long time, it is still necessary to use a convolution-like operator to convolve the input tokens to get a better representation. Character-level information also plays an important role for this task (13\% relatively improvement), but the performance would be heavily damaged if using characters only. 

\begin{table}
\centering
\scalebox{0.9}{
\begin{tabular}{ l|l|l|l}
\hline 
\bf Case & \bf Variant & \bf Dev & \bf Test \\ 
\hline \hline
\multirow{3}{*}{window size}
& $k=0$ & 93.96 & 94.06 \\ 
& $k=5$ & 94.27 & 94.81 \\ 
& $k=7$ & 94.52 & 94.71 \\ 
\hline
\multirow{4}{*}{character-level}
& character only & 92.17 & 93.0 \\
& $l_w = 0$ & 93.59 & 93.71 \\
& $l_w = 3$ & 94.21 & 94.41 \\
& $l_w = 7$ & 94.43 & 94.75 \\
\hline
\end{tabular}
}
\caption{Comparsion of hyper-parameters \label{tab:four}}
\end{table}

\subsection{Part-of-Speech Tagging}
Part-of-speech tagging is another sequence tagging task, which is to assign POS tags to each word in a sentence. It is very similar to the supertagging task. Therefore, these two tasks can be solved in a unified architecture. For POS tagging, we use the same network configurations as supertagging, except for the word vocabulary size and the tag set size. We conduct experiments on the Wall Street Journal of the Penn Treebank dataset, adopting the standard splits (sections 0-18 for the train, sections 19-21 for validation and sections 22-24 for testing). 

Although the POS tagging result presented in Table \ref{pos} is slightly below the state-of-the-art, we neither do any hyper-parameter tunings nor change the network architectures, just use the one getting the best test accuracy on the supertagging task. This proves the generalization of the model and avoids heavy work of model re-designing.

\begin{table}
\centering
\scalebox{0.9}{
\begin{tabular}{l|l}
\hline \bf Model & \bf Test \\ \hline \hline
S\o gaard \shortcite{sogaard2011semisupervised} & 97.5 \\
Ling et al. \shortcite{ling2015finding}  & 97.36 \\
Wang et al. \shortcite{wang2015part} & \bf 97.78 \\
Vaswani et al. \shortcite{vaswani2016supertagging} & 97.4 \\ 
Wu et al. \shortcite{wu2016skip} & 97.48\\
\hline
7-stacked mixed + non-linear gate & 97.48 \\
9-stacked mixed + non-linear gate & 97.53 \\
13-stacked mixed + non-linear gate & 97.51 \\
\hline
\end{tabular}
}
\caption{Accuracy for POS tagging on WSJ\label{pos}}
\end{table}

\section{Related Work}
Skip connections have been widely used for training deep neural networks. For recurrent neural networks, Schmidhuber \shortcite{schmidhuber1992learning}; El Hihi and Bengio \shortcite{el1995hierarchical} introduce deep RNNs by stacking hidden layers on top of each other. Raiko et al. \shortcite{raiko2012deep}; Graves \shortcite{graves2013generating};  Hermans and Schrauwen \shortcite{hermans2013training} propose the use of skip connections in stacked RNNs. However, the researchers have paid less attention to the analysis of various kinds of skip connections, which is our focus in this paper.

Recently, deep stacked networks have been widely used for applications. Srivastava et al. \shortcite{srivastava2015highway} and He et al. \shortcite{he2015deep} mainly focus on feed-forward neural network, using well-designed skip connections across different layers to make the information pass more easily. The Grid LSTM proposed by Kalchbrenner et al. \shortcite{kalchbrenner2015grid} extends the one dimensional LSTMs to many dimensional LSTMs, which provides a more general framework to construct deep LSTMs. 

Yao et al. \shortcite{yao2015depth} and Zhang et al. \shortcite{zhang2016highway} propose highway LSTMs by introducing gated direct connections between internal states in adjacent layers. Zilly et al. \shortcite{zilly2016recurrent} introduce recurrent highway networks (RHNs) which use a single recurrent layer to make RNN deep in a vertical direction.  These works do not use skip connections, and the hierarchical structure is reflected in the LSTM internal states or cell outputs. Wu et al. \shortcite{wu2016skip} propose a similar architecture for the shortcuts in stacked Bi-LSTMs. The difference is we propose generalized shortcut block architectures as basic units for constructing deep stacked models. We also discuss the compositions of these blocks.

There are also some works using stochastic gates to transmit the information. Zoneout \shortcite{krueger2016zoneout} provides a stochastic link between the previous hidden states and the current states, forcing the current states to maintain their previous values during the recurrent step. Chung et al. \shortcite{chung2016hierarchical} proposes a stochastic boundary state to update the internal states and cell outputs. These stochastic connections are connected between adjacent layers, while our constructions of the shortcuts are mostly cross-layered. Also, the updating mechanisms of LSTM blocks are different. 

\section{Conclusions}
In this paper, we propose the shortcut block as a basic architecture for constructing deep stacked models. We compare several gating functions and find that the non-linear deterministic gate performs the best. We also find the dense compositions perform better than the sparse ones. These explorations can help us to train deep stacked Bi-LSTMs successfully. Based on this shortcuts structure, we achieve the state-of-the-art results on CCG supertagging and comparable results on POS tagging. Our explorations could easily be applied to other sequence processing problems, which can be modeled with RNN architectures.


\bibliographystyle{tacl}
\bibliography{tacl}

\end{document}